\documentclass[journal,draftcls,onecolumn,12pt,twoside]{IEEEtran}

\usepackage{cite}
\usepackage{graphics}
\usepackage{amsmath,amsfonts, amssymb}
\usepackage{mathtools}
\usepackage{epstopdf}
\usepackage{amsthm}
\usepackage{bbm}
\usepackage{algorithm}
\usepackage[noend]{algpseudocode}
\usepackage{dsfont}
\usepackage{xcolor}
\usepackage{afterpage}
\usepackage{float}
\usepackage[justification=raggedright]{caption}
\usepackage{subcaption}

\DeclareMathOperator{\sign}{sign}

\ifCLASSINFOpdf
\else
\fi

\usepackage[colorinlistoftodos,textwidth=\marginparwidth]{todonotes}

\begin{document}
%
\title{ Time-Correlated Sparsification for Communication-Efficient Federated Learning}
%
%
%

\author{Mehmet Emre~Ozfatura,
       Kerem~Ozfatura ~and~
      Deniz~G{\"u}nd{\"u}z
\thanks{Emre Ozfatura and Deniz G{\"u}nd{\"u}z are with Information Processing and Communications Lab, Department of Electrical and Electronic Engineering,
Imperial College London Email: \{m.ozfatura, d.gunduz\} @imperial.ac.uk.}
\thanks{Kerem Ozfatura is with Department of Computer Science, Ozyegin University.} 
\thanks{This work was supported in part by the Marie Sklodowska-Curie Action SCAVENGE (grant agreement no. 675891), and by the European Research Council (ERC) Starting Grant BEACON (grant agreement no. 677854).}}

%
%
%

\maketitle

\thispagestyle{plain}
\pagestyle{plain}

\begin{abstract}
Federated learning (FL) enables multiple clients to collaboratively train a shared model without disclosing their local datasets. This is achieved by exchanging local model updates with the help of a parameter server (PS). However, due to the increasing size of the trained models, the communication load due to the iterative exchanges between the clients and the PS often becomes a bottleneck in the performance. Sparse communication is often employed to reduce the communication load, where only a small subset of the model updates are communicated from the clients to the PS. In this paper, we introduce a novel time-correlated sparsification (TCS) scheme, which builds upon the notion that sparse communication framework can be considered as identifying the most significant elements of the underlying model. Hence, TCS seeks a certain correlation between the sparse representations used at consecutive iterations in FL, so that the overhead due to encoding and transmission of the sparse representation can be significantly reduced without compromising the test accuracy. Through extensive simulations on the CIFAR-10 dataset, we show that TCS can achieve  centralized training accuracy  with $100$ times sparsification, and up to $2000$ times reduction in the communication load 
when employed together with quantization.

\end{abstract}


%
\IEEEpeerreviewmaketitle

\section{Introduction}
The success of  deep neural networks (DNN) in many complex machine learning problems has promoted their employment in a wide range of areas from finance \cite{NN.finance1} to healthcare \cite{FL.health2,FL.health3} and smart manufacturing \cite{Trakadas_2020}. However, one of the key challenges in utilizing DNNs in such applications is that often the training data is distributed across multiple institutions, and cannot be aggregated for centralized training due to the sensitivity of data \cite{FL.health1} and regulations\cite{FL.reg}. On the other hand, data available at in a single institution, such as a single bank, hospital, or a factory, may not be sufficient to train a ``sufficiently good" model with the desired generalization capabilities. Hence, collaborative training among multiple institutions/clients without sharing their local datasets addresses both the privacy concerns and the insufficiency of  local datasets. \cite{FL.health2,FL.health3,FL.health4}.

{\em Federated learning (FL)} framework has been introduced to address the aforementioned challenges in distributed machine learning \cite{FL1} by orchestrating the participating clients with the help of a central, so-called parameter server (PS), such that they can perform local training using their datasets first, and then seek a consensus on the global model by communicating with each other. They iterate between local training and consensus steps until converging to a global model. Such a strategy enables collaborative training without exchanging datasets; however, it requires exchange of a large number of parameter values over many iterations. The communication load can be a major bottleneck particularly when the underlying model is of high complexity, which increases the communication delays. For example, the VGG-16 and VGG-19 architectures have 138 and 144 million parameters \cite{NN.VGG}, respectively. Similarly, tens of millions of parameters are trained for speech recognition networks \cite{fohr:hal-01484447}. On top of that, in FL,  communication often takes place over bandwidth-limited channels \cite{FL_1, FL_2}. In general, communication becomes the bottleneck when the clients have relatively high computation speed compared to the throughput of the underlying communication network. To this end, communication-efficient designs are one of the key requirements for the successful implementation of FL over multiple clients in a federated manner.

\begin{figure*}[t]
\centering
\includegraphics[scale =1]{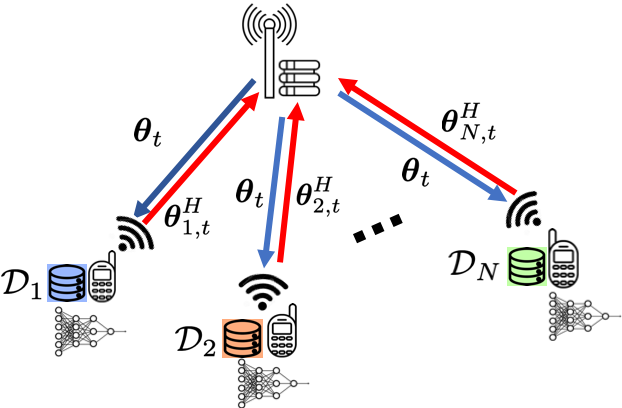}
\caption{Illustration of the FL system model and the FedAvg algorithm.}
\label{f:FL_model}
\end{figure*}

\subsection{Preliminaries}
The objective of  FL is to solve the following optimization problem over $N$ clients 
\begin{equation}
\min_{\boldsymbol{\theta}\in\mathbb{R}^{d}} f(\boldsymbol{\theta})= \frac{1}{N}\sum^{N}_{n=1}\underbrace{\mathds{E}_{\zeta_{n} \sim \mathcal{D}_{n}}f(\boldsymbol{\theta},\zeta_{n})}_{\mathrel{\mathop:}=f_{n}(\boldsymbol{\theta})},\label{DSO}
\end{equation}
where $\boldsymbol{\theta}\in\mathbb{R}^{d}$ denotes the model parameters,  $\zeta_{n}$ is a random data sample, $\mathcal{D}_{n}$ denotes the dataset of client $n$, and $f$ is the problem specific empirical loss function. At each iteration of FL,  each client aims to minimize its local loss function $f_{n}(\boldsymbol{\theta})$ using the {\em stochastic gradient descent} method. Then, the clients seek a consensus on the model with the help of the PS. The most widely used consensus strategy is to periodically average the locally optimized parameter models, which is referred to as {\em federated averaging (FedAvg)}. The FedAvg procedure is summarized in Algorithm \ref{alg:fl}. See Fig. \ref{f:FL_model} for an illustration of the FL model across $N$ clients each with local dataset.

\indent At the beginning of iteration $t$, each client pulls the current global parameter model $\boldsymbol{\theta}_t$ from the PS. In order to reduce the communication load, each client  performs $H$ local updates before the consensus step, as illustrated in Algorithm \ref{alg:fl} (lines 5-6), where 
\begin{equation}
\mathbf{g}^{\tau}_{n,t}=\nabla_{\boldsymbol{\theta}}f_{n}(\boldsymbol{\theta}^{\tau-1}_{n,t},\zeta_{n,\tau})
\end{equation}
is the gradient estimate of the $n$-th client at $\tau$-th local iteration based on the randomly sampled local data $\zeta_{n,\tau}$ and $\eta_{t}$ is the learning rate.

\begin{algorithm}[t!]
\caption{Federated Averaging (FedAvg) }\label{alg:fl}
\begin{algorithmic}[1]
\For{$t=1,2,\ldots$}
    \For{$n=1,\ldots,N$} in parallel
        \State Pull $\boldsymbol{\theta}_{t}$ from PS: $\boldsymbol{\theta}^{0}_{n,t}=\boldsymbol{\theta}_{t}$
        \For{$\tau=1,\ldots,H$}
            \State Compute SGD: $\mathbf{g}^{\tau}_{n,t}=\nabla_{\boldsymbol{\theta}}f_{n}(\boldsymbol{\theta}^{\tau-1}_{n,t},\zeta_{n,\tau})$
            \State Update model: $\boldsymbol{\theta}^{\tau}_{n,t}=\boldsymbol{\theta}^{\tau-1}_{n,t}-\eta_{t}g^{\tau}_{n,t}$
       \EndFor
       \State Push $\boldsymbol{\theta}^{H}_{n,t}$
    \EndFor
    \State{\textbf{Federated Averaging}:} $\boldsymbol{\theta}_{t+1}=\frac{1}{ \vert \mathcal{S}_{t} \vert}\sum_{n\in\mathcal{S}_{t}} \boldsymbol{\theta}^{H}_{n,t}$
\EndFor
\end{algorithmic}
\end{algorithm}

We note that when $H=1$, clients can send their local gradient estimates instead of updated models, and this particular implementation is called {federated SGD (FedSGD)}. For the sake of completeness, we also want to highlight that when the number of participating clients are large, e.g., FL across mobile devices, the PS can choose a subset of the clients  for the consensus to reduce the communication overhead. However, in the scope of this work, we consider a scenario with a moderate number of clients, all of which participate in all the iterations of the learning process. This would be the case when the clients represent institutions, e.g., hospitals or banks; and hence, client selection is not required. Similarly, in the case of  federated edge learning (FEEL) \cite{oa1, oa4, Gunduz:CL:20}, the number of colocated wireless devices participating in the training process may be  limited.

In addition to multiple local iterations, we can also employ compression of model updates in order to reduce the communication load from the clients to the PS at each iteration. Next, we briefly explain common compression strategies used in conveying the model updates from the clients to the PS in an efficient manner.
\subsection{Compressed Communication}
\indent The global model update in Algorithm \ref{alg:fl} (line 8) can be equivalently written in the following form:
\begin{equation} \label{model_update_mult}
\boldsymbol{\theta}_{t+1}= \boldsymbol{\theta}_{t}+\frac{1}{N}\sum^{N}_{n=1}\underbrace{\sum^{H}_{\tau=1}-\eta_{t}\mathbf{g}^{\tau}_{n,t}}_{\Delta\boldsymbol{\theta}_{n,t}}, 
\end{equation}
where we call the term $\Delta\boldsymbol{\theta}_{n,t}$ the model difference of the $n$th client at iteration $t$. Hence, each client can send the model difference instead of the updated model, and the compression is applied to this model difference. We can group the compression strategies into  two  main categories; namely, quantization and sparsification.

\subsubsection{Quantization}
In general, floating point precision with 32 bits is used for training DNNs, thus 32 bits are required to represent each element of the local gradient estimate. Quantization techniques aims to represent each element with fewer bits to reduce the communication load \cite{SGD.q0, SGD.q1, SGD.q2, SGD.q3, SGD.q4, SGD.q5, SGD.q6, Amiri:FedQuantPS}. In the most extreme case, only the sign of each element can be sent, i.e., using only a single bit per dimension, to achieve up to $\times32$ reduction in the communication load \cite{SGD.q0, SGD.q4, SGD.q5, SGD.q6}. Simulations with vision and speech recognition models show that significant reduction in the communication bandwidth can be achieved through quantization without much reduction in the network performance. 

\subsubsection{Sparsification} 
Sparsification techniques transform a $d$ dimensional vector of gradient estimate $\mathbf{g}$ to its sparse representation $\tilde{\mathbf{g}}$, where the non-zero elements of $\tilde{\mathbf{g}}$ are equal to the corresponding elements of $\mathbf{g}$. Sparsification can be considered as applying a $d$-dimensional mask vector $\mathbf{m}\in \left\{0,1\right\}^d$ on $\mathbf{g}$; that is, $\tilde{\mathbf{g}} = \mathbf{m}\otimes \mathbf{g}$, where $\otimes$ denotes element-wise multiplication. We denote the sparsification ratio by $\phi$, i.e.,
\begin{equation}
\phi \triangleq  \frac{\vert\vert\mathbf{m}\vert\vert_{1}}{d} << 1~.
\end{equation}
It has been shown that it is  possible to achieve sparsification ratios in the range of $\phi\in[0.01,0.001]$ for training dense DNN architectures, such as ResNet or VGG, without an apparent loss in test accuracy \cite{SGD.sparse1, SGD.sparse2, SGD.sparse3, SGD.sparse4, SGD.sparse5, SGD.sparse6, SGD.sparse7, SGD.sparse.rtopk, SGD.sparse.FL}.

We also would like to remark that the three strategies mentioned above, i.e., multiple local updates, quantization, and sparsification, are orthogonal to each other, can be employed together to further reduce the required communication bandwidth in FL.

For  collaborative/distributed learning, sparsification is commonly adopted in practice, and we identify its two popular variations in the literature: {\em top-$K$ sparsification} and {\em rand-K sparsification} \cite{SGD.sparse8}. Next, we briefly introduce the two approaches, and present their advantages and disadvantages in order to motivate the novel sparsification strategy we introduce in this work.

\begin{figure}
\begin{subfigure}{.5\textwidth}
\begin{center}
  \includegraphics[width=.9\linewidth]{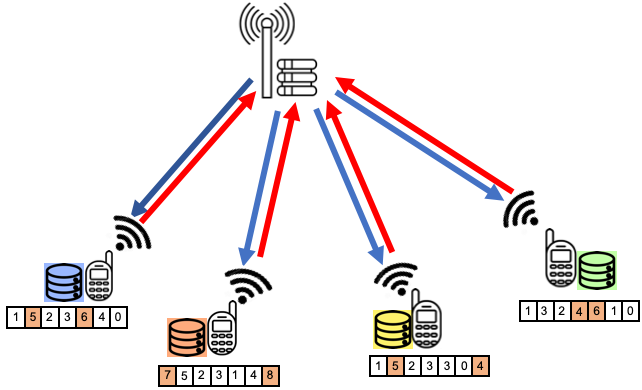}
  \caption{Top-$K$ sparsification with $K=2$.}
  \label{fig:topK}
\end{center}
\end{subfigure}%
\begin{subfigure}{.5\textwidth}
  \begin{center}
  \includegraphics[width=.9\linewidth]{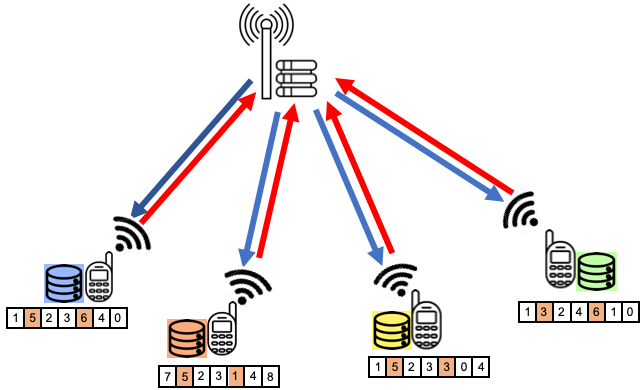}
  \caption{Rand-$K$ sparsification with $K=2$.}
  \label{fig:randK}
  \end{center}
\end{subfigure}
\caption{Illustration of the top-$K$ and rand-$K$ sparsification strategies in FL. Vectors at each client represent the model update in an iteration. The colored entries represent the entries that are sent to the PS.}
\label{fig:sparse}
\end{figure}

\subsection{Top-$K$ versus Rand-K Sparsification}

In top-$K$ sparsification, each client constructs its own mask $\mathbf{m}_{n,t}$ independently, based on the greatest absolute values in $\Delta\boldsymbol{\theta}_{n,t}$. See Fig. \ref{fig:topK} for an illustration of the top-$K$ sparsification strategy. Although top-$K$ sparsification is a promising compression strategy, its benefits in practical implementations are limited due to certain drawbacks. First, it requires a sorting operation which is computationally expensive when the size of the DNN architecture is large, which is the case when sparsification is most needed. Second, encoding and communication of the non-sparse locations induce additional overhead. Note that, when a client sends only $\phi d$ entries to the PS after sparsification, the compression ratio in terms of the communication bandwidth is less than $\phi$, since each client is required to send the locations of the all the non-zero parameters as well as their values for the PS to be able to recover the original update vector. More specifically, additional $\log_2 d$ bits must be transmitted to the PS to represent the location of each non-zero parameter. We want to emphasize that this overhead becomes more significant when the sparsification strategy is combined with quantization; that is, if $q<\log_2 d$ bits are used to represent the value of each non-zero parameter, then the communication of the non-zero positions becomes the main communication bottleneck. Finally, due to the mismatch between the masks of the clients, there might be a large gap between the sparsity in the uplink and downlink directions. We note that it has been shown in \cite{SGD.sparse.gtopk, SGD.sparse9} that it is possible to reduce the computational complexity of top-$K$ sparsification, hence in the scope of this paper we mainly focus on the last two drawbacks.\\
\indent On the other hand, in rand-K sparsification, the sparsity mask is constructed randomly at each iteration. However, by using the same seed, all the clients can use the same random sparsification mask $\mathbf{m}_{t}$. See Fig. \ref{fig:randK} for an illustration of the rand-$K$ sparsification strategy. Hence, one of the key advantages of rand-K compared to top-$K$, thanks to its pseudo-randomness, is that the clients do not need to encode and send the non-zero positions of their masks since they use an identical mask, which can also be known to the PS. In the case of distributed learning in a cloud center, this would allow using  more efficient communication protocols that requires linearity, such as the {\em allreduce} operation \cite{SGD.sparse8,sgd.sparse10}. Similarly, in the case of FEEL, this would allow employing over-the-air computation in an efficient manner \cite{oa1, oa2, oa3, oa4}. Besides, rand-K, particularly its block-wise implementation, is more efficient compared to top-$K$ both in terms of computation and memory access complexity \cite{SGD.sparse8, sgd.sparse10}. Despite all the aforementioned advantages, rand-K sparsification introduces larger compression errors compared to top-$K$; and thus, performs poorly in the high compression regime \cite{SGD.sparse.gtopk, SGD.sparse8}.\\
\indent In the light of the above discussions, we emphasize that the optimal sparsification strategy depends on the scenario. For instance, when we consider distributed learning within a cluster with sufficiently fast connection, low compression rates are sufficient to scale distributed computation efficiently \cite{PSGD.scale}, hence rand-K might be the right option \cite{SGD.sparse8}. However, in FL, where clients are geographically separated and communicate with the PS over low-speed links, top-$K$ might be the only option.

\subsection{Sparse Network Architectures: Connections to Dynamic Network Pruning}

{\em Network pruning}  aims to reduce the size and complexity of DNNs \cite{NN.prune1,NN.prune2,NN.prune3,NN.prune4,NN.prune5,NN.prune6}. Given an initial DNN model $\boldsymbol{\theta}$, the objective of network pruning is to find a sparse version of $\boldsymbol{\theta}$, denoted by $\widetilde{\boldsymbol{\theta}}$, where only a small subset of the parameters are utilized with minimal loss in accuracy. In other words, the objective is to construct a $d$-dimensional mask vector $\mathbf{m}_p\in \left\{0,1\right\}^d$ to recover $\widetilde{\boldsymbol{\theta}} =\mathbf{m}_p\otimes\boldsymbol{\theta}$, where $\vert\vert\mathbf{m}_p\vert\vert_{1}<<d$. The `lottery ticket hypothesis' \cite{NN.prune6} further states that such a mask can be employed throughout training to obtain the same level of accuracy with similar training time. We note that the existence of a ``good" mask with $\vert\vert\mathbf{m}_p\vert\vert_{1}<<d$ and a similar test accuracy as the unpruned network implies the existence of a sparse communication strategy during training. 
If the optimal pruning mask $\mathbf{m}_p$ is used throughout training, this would also result in a significant reduction in the communication load by more than a factor of $\phi = \vert\vert\mathbf{m}_p\vert\vert_{1} / d$, as the clients  need to convey only $\vert\vert\mathbf{m}_p\vert\vert_{1}$ values, and there is no need to specify their locations. However, the optimal pruning mask typically cannot be determined at the beginning of the training process. Alternatively, in \textit{dynamic network pruning} \cite{NN.prune1,NN.prune2,NN.prune3,NN.prune4,NN.prune5}, an evolving sequence of masks are employed over iterations, where the pruning mask employed at each iteration is updated gradually.

\subsection{Motivation and Contributions}

At each iteration of the top-$K$ sparsification strategy, each client constructs a mask vector $\mathbf{m}_{n,t}$ from scratch, independently of the previous iterations, although the gradient values, and hence, the top-$K$ positions, exhibit certain correlation over time. The core idea of our work is to exploit this correlation  to reduce the communication load resulting from the transmission of non-zero parameter locations. In other words, if the {\em significant } locations, those often observe  large gradient values, were known,  the clients could simply communicate the values corresponding to these locations without searching for the top-$K$ locations, or specifying their indices to the PS, reducing significantly  both the communication load and the complexity.

More specifically, inspired by the dynamic pruning techniques \cite{NN.prune1, NN.prune2, NN.prune3, NN.prune4, NN.prune5}, we propose a novel sparse communication strategy, called time correlated sparsification (TCS), where we search for a ``good" global mask $\mathbf{m}_{t}$ to be used by all the clients at iteration $t$, which evolves gradually over iterations. Client $n$ uses a slightly personalized mask $\mathbf{m}_{n,t}$ based on the global mask $\mathbf{m}_{t}$, where
\begin{equation}\label{eqn:tcs}
\vert\vert  \mathbf{m}_{n,t}-\mathbf{m}_{t}  \vert\vert_1 = \epsilon_t << \phi d.
\end{equation}
\indent This correlation between $\mathbf{m}_{n,t}$ and  $\mathbf{m}_{t}$ helps to reduce the communication load since $\mathbf{m}_{t}$ is known at the PS. The small variations serve two purposes: First,  they are used to `explore' a small portion of new locations to improve the current mask. Second, certain locations may become significant {\em temporarily}  due to the  accumulation of  errors when an error feedback mechanism is employed \cite{SGD.sparse2, SGD.feedback2, SGD.feedback3, SGD.feedback_double1, SGD.feedback_double2}.

TCS  is designed to benefit from the strong aspects of  both the rand-K and top-$K$ sparsification strategies. The advantages of the proposed TCS strategy can be summarized as follows:
\begin{itemize}
\item Compared to top-$K$ sparsification, each client encodes and sends only a small number of non-sparse positions at each iteration, which reduces the number of transmitted bits. Furthermore, it makes high compression rates possible when TCS is combined with  quantization.
\item Although both TCS and top-$K$ offers the same sparsification level in the uplink direction, TCS achieves much higher sparsification level in the downlink direction thanks to the correlation of the masks across clients.
\item Finally, since the individual masks mostly coincide, in a FEEL scenario, where the clients communicate with the PS over a shared wireless channel, TCS allows employing over-the-air computation in an efficient manner and takes advantage of the superposition property of the  wireless medium \cite{oa1, oa2, oa3, oa4}. 
\end{itemize}

Through extensive simulations on the CIFAR-10 dataset, we show that TCS can achieve centralized training accuracy with $100$ times sparsification, and up to $2000$ times reduction in the communication load when employed together with quantization. We also note that, in the scope of this paper, we consider the general setup without any specification of the communication medium, and will consider the FEEL scenario with over-the-air computation in an extension of this work.




\section{Time Correlated Sparsification (TCS)}
\subsection{Design principle}
\indent The main design principle behind TCS is employing two distinct mask vectors for sparsification. $\mathbf{m}_{global}$ is used to exploit previously identified important DNN parameters, whereas $\mathbf{m}_{local}$ explores new parameters. In particular, $\mathbf{m}_{global}$ is constructed based on the previous model update $\Delta{\boldsymbol{\theta}}$, motivated by the assumption that the most important DNN weights do not change significantly over iterations. Since all the clients receive the same global model update from the PS, mask vector $\mathbf{m}_{global}$ is identical for all the clients.\\
\indent Let $\phi_{global}$ be the sparsification ratio for the mask $\mathbf{m}_{global}$ such that
\begin{equation}
\vert\vert\mathbf{m}_{global}\vert\vert_{1}=K_{global}=\phi_{global}\cdot d.
\end{equation}
At the beginning of iteration $t$, each client receives the global model difference $\Delta\boldsymbol{\theta}_{t-1}$ from the PS, and accordingly, obtains $\mathbf{m}_{global}$ by simply identifying the top $K_{global}$ values in $\vert \Delta\boldsymbol{\theta}_{t-1}\vert$. In parallel, each client uses the received global model difference to recover $\boldsymbol{\theta}_{t}$.\\
\indent Following the model update, each client $n\in[N]$ carries out the local SGD steps, and  
computes the local model difference $\Delta\boldsymbol{\theta}_{n,t}$. Finally, it obtains the sparse version of the local model difference, $\hat{\Delta}\boldsymbol{\theta}_{n,t}$, using the mask vector $\mathbf{m}_{global}$, i.e.,
\begin{equation}
\hat{\Delta}\boldsymbol{\theta}_{n,t} =  \mathbf{m}_{global} \otimes \Delta\boldsymbol{\theta}_{n,t}.
\end{equation}
We want to emphasize that since the PS sent out $\Delta{\boldsymbol{\theta}}_{t-1}$, mask vector $\mathbf{m}_{global}$ is known by the PS as well. Therefore, for each client it is sufficient to send only the non-zero values of $\hat{\Delta}\boldsymbol{\theta}_{n,t}$, without specifying their positions. Therefore, compared to the conventional top-$K$ sparsification framework, TCS further reduces the communication load by removing the need to communicate the positions of the non-zero values.\\
\indent However, the main drawback of the above approach is that, if $\mathbf{m}_{global}$ is used throughout the training process, the same subset of weights will be used for model update at all iterations. Therefore, the proposed sparsification strategy requires a feedback mechanism in order to explore new weights at each iteration to check whether there are more important weights to consider for model update.\\
\indent To introduce such a feedback mechanism, each client $n$ employs a second mask $\mathbf{m}^{n}_{local}$, which is unique to that client. The feedback mechanism works in the following way: given $\hat{\Delta}\boldsymbol{\theta}_{n,t}$, $\mathbf{m}^{n}_{local}$ is obtained as the vector of the greatest $K_{local}=\phi_{local}\cdot d$ entries of $\vert\Delta\boldsymbol{\theta}_{n,t}-\hat{\Delta}\boldsymbol{\theta}_{n,t}\vert$. Hence, at each iteration $t$, client $n$ sends, $n \in [N]$, 
\begin{equation}
 	\widetilde{\Delta}\boldsymbol{\theta}_{n,t}=\Delta\boldsymbol{\theta}_{n,t}\otimes(\mathbf{m}_{global}+\mathbf{m}^{n}_{local})
\end{equation}
to the PS.\\
\indent Since the main purpose of $\mathbf{m}^{n}_{local}$ is to explore new important parameters, we assume that $\phi_{local}<<\phi_{global}$. Assume that $q$ bits are used to represent the value of each parameter. Then, using the global sparsification mask, the total number of bits to be conveyed to the PS is $K_{global} \cdot q$. For the feedback mechanism, in addition to $q$ bits to represent each of the parameter values, $\log_2 d$ bits are required to inform the PS about each position of the parameter within the $d$-dimensional update vector. Hence, the total number of bits transmitted at each iteration is given by:
\begin{equation}
 	Q_{TCS}= q \cdot d \cdot (\phi_{local}+\phi_{global})+ \log_2 d \cdot d \cdot \phi_{local}.
\end{equation}

Here, we would like to note that by utilizing a more efficient encoding strategy, it is possible to represent each position with $\log_2(1/\phi_{local})+2$ bits, instead of $\log_2 d$, which is more communication efficient when $d$ is large. We refer the reader to Subsection \ref{sec:TCS-se} for the details of this encoding strategy. We  want to emphasize that since $\phi_{local}<<\phi_{global}$, the proposed TCS strategy is more communication efficient than top-$K$ sparsification with the same $\phi_{global}$ value, such that $K = \phi_{global} \times d$. The total number of required bits are given as 
\begin{equation}
 	Q_{topK}=  d \cdot \phi_{global} \cdot (q + \log_2 d).
\end{equation}
One can easily observe that for $\phi_{local}<<\phi_{global}$, $Q_{TCS}$ is smaller than $Q_{topK}$, especially when $q$ is small.
\subsection{Error accumulation}
Due to sparsification, there is an error in the local model difference sent to the PS by client $n$, which can be expressed as:
\begin{equation}
\mathbf{e}_{n,t}=\Delta\boldsymbol{\theta}_{n,t}\otimes(1-\mathbf{m}_{global}-\mathbf{m}^{n}_{local}).
\end{equation}
It has been shown that the convergence speed can be improved  by propagating the current compression error to next iterations \cite{SGD.feedback1, SGD.feedback2, SGD.feedback3}. That is, at iteration $t$, client $n$ intends to send 
\begin{equation}
\bar{\Delta}\boldsymbol{\theta}_{n,t}=\Delta\boldsymbol{\theta}_{n,t}+\mathbf{e}_{n,t-1}
\end{equation}
to the PS. Accordingly, each client performs sparsification on $\bar{\Delta}{\boldsymbol{\theta}}_{n,t}$ instead of $\Delta{\boldsymbol{\theta}}_{n,t}$. The overall TCS algorithm with error accumulation is summarized in Algorithm \ref{alg:time_corr}.

\begin{algorithm}[t]
\caption{TCS with error accumulation}\label{alg:time_corr}
\begin{algorithmic}[1]
    \For{$t=1,\ldots,T$}
    \State\textbf{Client side:}
     \For{$n=1,\ldots,N$} in parallel
     \State Receive $\Delta\boldsymbol{\boldsymbol{\theta}}_{t-1}$ from PS
     \State $\mathbf{m}_{global}=S_{top}(\Delta\boldsymbol{\theta}_{t-1},K_{global})$
     \State \textbf{Update model:} $\boldsymbol{\boldsymbol{\theta}}_{n,t} = \boldsymbol{\theta}_{n,t-1} + \Delta\boldsymbol{\theta}_{t-1}$
    \State Perform $H$ local updates and compute $\Delta{\boldsymbol{\theta}}_{n,t}$ 
    \State \textbf{Error Feedback:}
    \State $\bar{\Delta}{\boldsymbol{\theta}}_{n,t}=\Delta{\boldsymbol{\theta}}_{n,t}+\mathbf{e}_{n,t-1}$
    \State $\mathbf{m}^{n}_{local}=S_{top}(\bar{\Delta}{\boldsymbol{\theta}}_{n,t}\otimes(1-\mathbf{m}_{global}),K_{local})$
    \State $\widetilde{\Delta}\boldsymbol{\boldsymbol{\theta}}_{n,t}=(\mathbf{m}^{n}_{local} + \mathbf{m}_{global})\otimes\bar{\Delta}{\boldsymbol{\theta}}_{n,t}$
    \State Send $\widetilde{\Delta}\boldsymbol{\theta}_{n,t}$ to PS
    \State $\mathbf{e}_{n,t}= \bar{\Delta}\boldsymbol{\theta}_{n,t}-\widetilde{\Delta}\boldsymbol{\theta}_{n,t}$
    \EndFor
\State{Aggregate local model differences:}
\State $\Delta\boldsymbol{\theta}_t=\frac{1}{N}\sum_{n\in[N]}\widetilde{\Delta}\boldsymbol{\theta}_{n,t}$
\State Send $\Delta\boldsymbol{\theta}_t$ to clients
    \EndFor
\end{algorithmic}
\end{algorithm}
We note that $S_{top}(\mathbf{v},K)$ in Algorithm  \ref{alg:time_corr} maps  vector $\mathbf{v}\in \mathbb{R}^d$ to a mask vector  $\mathbf{m}\in \left\{0,1\right\}^d$ such that if $\overline{v}_{K}$ is the $K$th greatest value in $\vert\mathbf{v}\vert$, then
\begin{equation}
\mathbf{m}_{i}= 1 ~ \text{if} ~\vert\mathbf{v}_{i}\vert \geq \overline{v}_{K},
\end{equation}
and $0$ otherwise.

\subsection{Layer-wise fairness}
Due to the layered structure of the DNNs and the backpropagation mechanism, gradient values do not have uniform distribution across the layers. As a consequence, when top-$K$ sparsification is employed, the gradient of a weight belonging to initial layers is more likely to be chosen in the sparsified gradient. In other words, gradient values corresponding to later layers will be discarded more often, which may affect the final performance. Furthermore, at each iteration, by using a local mask vector $\mathbf{m}_{local}$, each client suggests $K_{local}$ new parameters to the PS to consider for sparsification, but again, we expect these locations to be distributed in a non-uniform manner across the DNN layers. To mitigate this problem, we introduce a layer-wise fairness constraint which introduces a distinct maximum sparsification ratio for each DNN layer in addition to the given sparsification parameters. The resultant scheme is called TCS with layer-wise fairness (TCS-LF).\\

\indent Let the DNN architecture consist of $L$ layers, with $d_l$ parameters in the $l$-th layer, $l \in [L]$. The mask vectors are formed by concatenating $L$ mask vectors, each corresponding to a different layer. Let $\phi^{max}_{local}$ and $\phi^{max}_{global}$ denote the maximum sparsification levels allowed for the construction of the mask vectors $\mathbf{m}_{local}$ and $\mathbf{m}_{global}$, respectively. Similarly, let $\mathbf{m}^{l}_{local}$ and $\mathbf{m}^{l}_{global}$ denote the local and global mask vectors, respectively, for the $l$-th layer, $l \in [L]$. Then, the layer-wise fairness requiries 
\begin{equation}
\vert\vert\mathbf{m}^{l}_{global}\vert\vert_{1}\geq\phi^{max}_{global}\cdot d_{l},~ \forall l\in[L]
\end{equation}
and 
\begin{equation}
\vert\vert\mathbf{m}^{n,l}_{local}\vert\vert_{1}\geq\phi^{max}_{local}\cdot d_{l}, ~ \forall l\in[L], ~ \forall n\in[N].
\end{equation}
 TCS-LF  imposes a layer-wise sparsity constraint for both the clients ($\phi^{min}_{local}$) and the PS side ($\phi^{min}_{global}$) by including parameters from every individual layer of the network model. However, it is possible to consider fairness only at the PS side by only imposing the constraint $\phi^{min}_{global}$, which we call TCS with partial layer-wise fairness (TCS-PLF).

\subsection{TCS with momentum}
\label{sec:TCS-m}
Momentum SGD is a popular acceleration strategy used in training  DNNs, and it increases the convergence speed and provides better generalization \cite{SGD.opt}. Here, we illustrate how
momentum SGD optimizer can be  incorporated into the FedSGD framework with the proposed sparsification strategy. We note that, momentum SGD can also be utilized when clients perform multiple local iterations \cite{FL.acc1,FL.acc2}; however, in the scope of this paper we limit our focus to FedSGD with momentum. In FedSGD with sparsification,  each client sends sparsified gradient vector, $\widetilde{\mathbf{g}}_{n,t}$, and receives the sum of the sparsified gradients  $\widetilde{\mathbf{g}}_{t}$ from the PS. When the momentum SGD is used for the model update, each user first updates the momentum term $\mathbf{w}_{n,t}$ as follows:
\begin{equation}
\mathbf{w}_{n,t}=\beta \cdot\mathbf{w}_{n,t-1}+ \widetilde{\mathbf{g}}_{t-1},
\end{equation}
where $\beta$ is the momentum coefficient. Following the update of the momentum term, the local model is updated using the momentum:
\begin{equation}
\boldsymbol{\theta}_{n,t} = \boldsymbol{\theta}_{n,t-1} + \eta_{t} \cdot \mathbf{w}_{n,t}.
\end{equation}
We refer to the momentum term, $\mathbf{w}_{n,t}$, introduced above as the {\em global momentum}, since it is  identical for all the clients. The overall TCS framework with global momentum is summarized in Algorithm \ref{alg:time_corr_momentum}. 

\begin{algorithm}
\caption{TCS with global momentum  }\label{alg:time_corr_momentum}
\begin{algorithmic}[1]
    \For{$t=1,\ldots,T$}
    \State \textbf{Client side:}
     \For{$n=1,\ldots,N$} in parallel
     \State Receive $\tilde{\mathbf{g}}_{t-1}$ from PS
     \State $\mathbf{m}_{global}=S_{top}(\mathbf{g}_{t-1},K_{global})$
     \State \textbf{Update momentum:}
     \State{$\mathbf{w}_{n,t}=\mathbf{w}_{n,t-1}+\beta\bar{\mathbf{g}}_{t-1}$}
     \State \textbf{Update model:}
     $\boldsymbol{\theta}_{n,t} = \boldsymbol{\theta}_{n,t-1} - \eta_{t} \cdot \mathbf{w}_{n,t}$
    \State \textbf{Compute SGD:} $\mathbf{g}_{n,t}=\nabla_{\boldsymbol{\theta}}F(\boldsymbol{\theta}_{n,t},\zeta_{n,t})$
    \State $\bar{\mathbf{g}}_{n,t}=\mathbf{g}_{n,t}+\mathbf{e}_{n,t-1}$
    \State $\mathbf{m}^{n}_{local}=S_{top}(\bar{\mathbf{g}}_{n,t}\otimes(1-\mathbf{m}_{global}),K_{local})$
    \State $\widetilde{\mathbf{g}}_{n,t}=(\mathbf{m}^{n}_{local} + \mathbf{m}_{global})\otimes\bar{\mathbf{g}}_{n,t}$
    \State Send $\widetilde{\mathbf{g}}_{n,t}$ to PS
    \State $\mathbf{e}_{n,t}= \bar{\mathbf{g}}_{n,t}-\widetilde{\mathbf{g}}_{n,t}$
    \EndFor
\textbf{PS side:}
\State{Aggregate local gradients:}
\State $\tilde{\mathbf{g}}_t=\frac{1}{N}\sum_{n\in[N]}\widetilde{\mathbf{g}}_{n,t}$
\State Send $\tilde{\mathbf{g}}_t$ to clients
    \EndFor
\end{algorithmic}
\end{algorithm}

\subsection{Sparsity encoding}\label{sec:TCS-se}
In this subsection, the position of the non-zero values. In general, $\log_2 d$ bits are sufficient to send the position of each non-zero element of the vectors to be conveyed. However, for the local sparsification step based on $\mathbf{m}^{n}_{local}$, we employ the following coding scheme to send the position of the non-zero values. Let $\mathbf{v}$ be a sparse vector, where $\phi$ portion of its indices are non-zero. Initially, we assume $\mathbf{v}$ is divided into equal-length blocks of size $1/\phi$. The position of each non-zero value in $\mathbf{v}$ can be defined by two parameters; namely, {\em BlockIndex} and {\em IntraBlockPosition} which denotes the corresponding block and its position within the block, respectively. Hence, {\em IntraBlockPosition} can be represented with $\log_2 (1/\phi)$ bits. To specify the {\em BlockIndex} two additional bits are sufficient. We use $0$ to identify the end of each block, and append $1$ to the beginning of the $\log_2 (1/\phi)$ bits used for {\em IntraBlockPosition}. Hence, on average $\log_2 (1/\phi) + 2$ bits are needed to represent the location of each non-zero value.\\
\indent In the decoding part, given the encoded binary vector $\mathbf{v}_{loc}$ for the position of the non-zero values in $\mathbf{v}$, the PS starts reading the bits from the first index and checks whether it is $0$ or $1$. If it is $1$, then the next $\log_2 (1/\phi)$ bits are used to recover the position of the non-zero value in the current block. If the index is $0$, then the PS increases {\em BlockIndex}, which tracks the current block, by one, and moves on to the next index. The overall decoding procedure is illustrated in Algorithm \ref{alg:encode}.
\begin{algorithm}
\caption{Sparse position decoding}\label{alg:encode}
\begin{algorithmic}[1]
\State{Input: Encoded sparse position vector $\mathbf{v}_{loc}$}
\State{Initialize: $pointer=0$}
\State{Initialize: $BlockIndex$}
\While{$pointer<length(\mathbf{v}_{loc})$}
\If{$\mathbf{v}_{loc}(pointer)=0$}
\State{$BlockIndex=BlockIndex+1$}
\State{$pointer=pointer+1$}
\Else
\State{Read  next $\log_2(1/\phi)$ bits for $IntraBlockPosition$}
\State{Recover the location of a non-zero value:}
\State{$(1/\phi)\cdot BlockIndex+IntraBlockPosition$}
\State{$pointer=pointer + \log_2 (1/\phi) +1$}
\EndIf
\EndWhile
\end{algorithmic}
\end{algorithm}

To elucidate the sparse encoding, consider a sparse vector $\mathbf{v}$ of size $d=12$ with $\phi=1/4$ and the indices of the non-zero values are given as  $1, 3, 10$. Since $\phi=1/4$, the vector is divided into 3 blocks, each of size $4$ and  {\em IntraBlockPosition} of each non-zero value  can be represented by 2 bits, i.e., $00,10,01$; hence, overall the sparse representation can be written as a binary vector of
\begin{equation}
[{\color{green}1}{\color{red}00}{\color{green}1}{\color{red}10}{\color{blue}0}{\color{blue}0}{\color{green}1}{\color{red}01}{\color{blue}0}],
\end{equation}
where the bits in red represents the position within the block, bits in green indicate that the following two bits refer to the position within the current block, and bits in blue represent the end of a block.

\subsection{Fractional quantization}
There is already a rich literature on quantized communication in the distributed SGD framework \cite{SGD.q0, SGD.q1, SGD.q2, SGD.q3, SGD.q4, SGD.q5, SGD.q6}. Although the most common approaches are TernGrad \cite{SGD.q2} and QSGD  \cite{SGD.q2}, we consider the scaled sign operator \cite{SGD.feedback2,SGD.q0} for quantization. For a given $d$ dimensional vector $\mathbf{u}$, scaled sign operator maps the value of $i$th parameter, $\mathbf{u}_{i}$, to a quantized scalar value, $\mathcal{Q}(\mathbf{u}_{i})$, in the following way:
\begin{equation}
\mathcal{Q}(\mathbf{u}_{i})=\frac{\vert\vert \mathbf{u}\vert\vert_{1}}{d}\sign(\mathbf{u}_{i}),
\end{equation}
where $\sign(\cdot)$ is the sign operator. It has been shown that the impact of the quantization error can be reduced by dividing $\mathbf{u}$ into $P$ smaller disjoint blocks $\left\{\mathbf{u}_{1},\ldots, \mathbf{u}_{P} \right\}$, and then applying quantization to each block separately, and often these blocks correspond to layers of DNN architecture \cite{SGD.q0}. Although we follow the same approach, inspired by the natural compression approach in \cite{SGD.q9}, we utilize a different quantization scheme. Let $u_{max}$ and $u_{min}$ be the maximum and minimum values in vector $\vert\mathbf{u}\vert$, respectively. We divide the interval $[u_{max},u_{min}]$ into $P$ disjoint intervals $I_{1},\ldots,I_{P}$, such that the $p$th interval, $I_{p}$, is given as
\begin{equation}
I_{p}=[\sigma^{p-1}u_{max},\sigma^{p}u_{max}]
\end{equation}
where,
\begin{equation}
\sigma=\left(\frac{u_{min}}{u_{max}}\right)^{1/P}.
\end{equation}
Further, let $\mu_{p}$ be the average of the values assigned to interval $I_{p}$. Then fractional quantization maps the value of $i$th parameter, $\mathbf{u}_{i}$, to a quantized scalar value, $\mathcal{Q}_{f}(\mathbf{u}_{i})$, in the following way:
\begin{equation}
\mathcal{Q}_{f}(\mathbf{u}_{i})=\sum^{P}_{p=1}\mathds{1}_{\left\{\mathbf{u}_{i}\in I_{p}\right\}}\mu_{p}\sign(\mathbf{u}_{i}),
\end{equation}
where $\mathds{1}_{\left\{\cdot\right\}}$ is the indicator function. Since there are $P$ intervals in total, $\log_2 P$ bits are sufficient to identify  the corresponding interval of $\mathbf{u}_{i}$, $i=1,\ldots,d$. An  additional bit is sufficient to represent the sign, hence in total $\log_2 P +1$ bit are required per parameter. Additional $32 \cdot P$ bits are required to convey the mean values of the intervals. Consequently, for a given $d$ dimensional vector $\mathbf{u}$, a total of $d (\log_2 P + 1) + 32 \cdot P$ bits are required, and often the second term is negligible.\\
\indent One can observe that fractional quantization strategy ensures the following inequality for each parameter value $\mathbf{u}_{i}$:
\begin{equation}\label{comperror_loc}
\vert\mathcal{Q}_{f}(\mathbf{u}_{i})-\mathbf{u}_{i}\vert \leq \gamma  \vert\mathbf{u}_{i}\vert,
\end{equation}
where $\gamma=\frac{1-\sigma}{\sigma}$. Accordingly, similar inequality  holds for vector $\mathbf{u}$ and its quantized version $\tilde{\mathbf{u}}$ i.e.,
\begin{equation}\label{comperror}
\vert\vert\tilde{\mathbf{u}}-\mathbf{u}_{i}\vert\vert_{2} \leq \gamma  \vert\vert \mathbf{u}_{i}\vert\vert_{2}.
\end{equation}
We remark that inequality (\ref{comperror}) is often used to show the boundedness of the compression error, and to prove the convergence of distributed training with compression error \cite{SGD.feedback2}. Besides, as illustrated in inequality (\ref{comperror_loc}), fractional quantization also implies a proportional error for each parameter value, which provides a certain fairness among the parameters belonging to different layers of the DNN.

\section{Numerical Results}\label{sec:NR}
In this section, we provide numerical results compering all the introduced schemes above, and illustrate the remarkable improvements in the communication efficiency provided by TCS and its variants.

\subsection{Simulation Setup}
\indent To evaluate the performance of the proposed TCS strategy, we consider the image classification task on the CIFAR-10 dataset \cite{cifar-10}, which consists of 10 image classes, organized into 50K training and 10K test images, respectively. We employ the ResNet-18 architecture as the DNN \cite{NN.DRN}, which consists of 8 basic blocks, each with two 3x3 convolutional layers and batch normalization. After two consecutive basic blocks, image size is halved with an additional 3x3 convolutional layer. This network consists of 11,173,962 trainable parameters altogether. We consider a network of $N=10$ clients and a federated setup, in which the training dataset is divided among the clients in a disjoint manner. The images, based on their classes, are distributed in an identically and independently distributed (IID) manner among the clients.

\subsection{Implementation}
\indent For performance evaluation, we consider the centralized training as our main benchmark, where we assume that all the training dataset is collected at one client. We set the batch size to 128 and the learning rate to $\eta=0.1$. The performance of this centralized setting will be referred to as the Baseline in our simulation results. For all the FL strategies considered in this work we set the batch size to 64, and adopt the linear learning rate scaling rule in \cite{large_scale_training}, where the learning rate is scaled according to the cumulative batch size and the total number of samples trained by all the clients, taking the batch size of 128 as a reference value with the corresponding learning rate $\eta=0.1$. Hence, for our setup with $N=10$ clients, we use the learning rate $\eta=0.5$. Further, in all the FL implementations we employ the warm up strategy \cite{large_scale_training}, where the learning rate is initially set to $\eta=0.1$, and is increased to its corresponding scaled value gradually in the first 5 epochs. We also note that during the warm up phase we do not employ sparsification and quantization methods for communication.\\
\indent The DNN architecture is trained for 300 epochs and the learning rate is reduced by a factor of 10 after the first 150 and 225 epochs, respectively \cite{NN.DRL, NN.DRN}. Lastly, in all the simulations we employ L2 regularization with a given weight decay parameter $10^{-4}$.\\
\indent For performance evaluation, we consider the top-$K$ sparsification scheme as a second benchmark. For top-$K$ sparsification we set the sparsification ratio to $\phi = 10^{-2}$. Accordingly, for the proposed TCS strategy we set $\phi_{global} = 10^{-2}$ and $\phi_{local} = 10^{-3}$. We want to emphasize that for the TCS strategy with 10 clients, these parameters imply a maximum of $0.02$ sparsification ratio; in other words $\times50$ compression in the PS-to-client direction as well, which is not the case for top-$K$ sparsification.
For TCS-LF, we set $\phi^{min}_{local} = 4 \times 10^{-4}$ and $\phi^{min}_{global} = 10^{-3}$ for the network layers and consider $\phi^{min}_{global} = 10^{-3}$ for TCS-PFL as well.\\
\indent We recall that one of the key design parameters of FL is the number of local steps $H$. Hence, we use TCS-L$H$ to denote the TCS scheme with $H$ local iterations and use TCS to refer to the FedSGD scheme where $H=1$. For FedAvg, we consider $H=2$ and $H=2$ in our simulations.\\
\indent Finally, we also employ quantization strategy to represent each non-zero value with $q<<32$ bits and use the notation `Q$q$' to denote the number of bits used to represent each element.For example, TCS-L4-Q5 denotes to $H=4$ along with 5-bit quantization. For performance evaluation, we employ two performance metrics:  {\em test accuracy} and the {\em bit budget}, corresponding to the performance of the final trained model and the communication load, respectively. More specifically, the bit budget refers to the average number of bits conveyed from a client to the PS per parameter per iteration.

\subsection{Simulation Results}
\begin{figure*}[t]
\begin{center}
\includegraphics[scale =1.0]{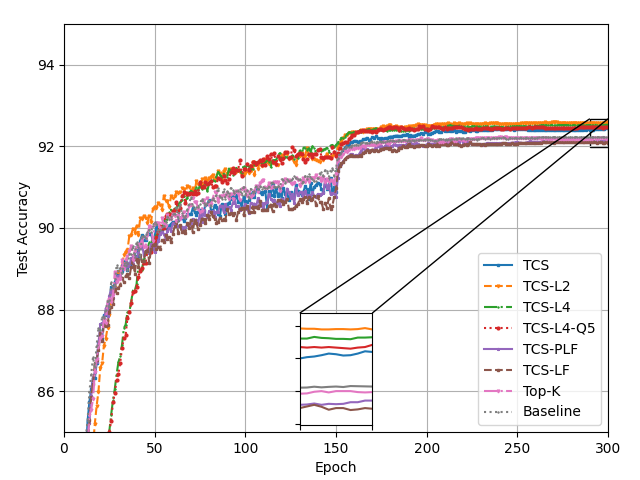}
\end{center}
\caption{Comparison of the test accuracy results for different FL techniques and the centralized baseline over 300 epochs (for $\eta=0.5$).}
\label{300epoch_05}
\end{figure*}

In our first simulation, we consider 8 schemes, namely the Baseline, top-$K$ sparsification, TCS, TCS-PLF, TCS-LF, TCS-L2, TCS-L4, and TCS-L4-Q5, where the first two are used as benchmark schemes. For each scheme we take the average over 5 trials. The final test accuracy results, with mean and standard deviation, and the bit budget for each scheme is presented in Table \ref{top1_acc_05}. In Figure \ref{300epoch_05}, we present the test accuracy results with respect to the epoch index.

\begin{table}
\begin{center}
\begin{tabular}{ |c | c| c|}
    \hline
    \textbf{Method} & \textbf{Test Accuracy (mean $\pm$ std)} & \textbf{Bit budget}\\ 
    \hline
    TCS & 92.44 $\pm$ 0.143 & 0.363\\
    \hline
    TCS-L2 &\textbf{92.578 $\pm$ 0.189}& 0.1815\\
    \hline
    TCS-L4 & 92.53 $\pm$ 0.22 & 0.0907\\
    \hline
    TCS-L4-Q5 & 92.485 $\pm$ 0.22 & 0.01675\\
    \hline
    TCS-PLF &92.142 $\pm$ 0.166 & 0.363\\
    \hline
    TCS-LF & 92.115 $\pm$ 0.324 & 0.363\\
    \hline
    top-$K$ & 92.194 $\pm$ 0.247 & 0.41\\
    \hline
    Baseline & 92.228 $\pm$ 0.232 & -\\
    \hline
\end{tabular}
\end{center}
 \caption{Test accuracy (for $\eta=0.5$) and bit budget of the studied schemes.}
 \label{top1_acc_05}
\end{table}

We observe that the proposed TCS scheme requires $12\%$ lower bit budget than top-$K$ sparsification while achieving a higher average test accuracy. Similarly, it achieves approximately $\times100$ reduction in the communication load without losing the accuracy. 
We also observe that TCS with multiple local iterations, in particular TCS-L2 and TCS-L4, achieve higher test accuracy compared to TCS with single local iteration. While the best accuracy achieved by TCS-L2, TCS-L4 achieves almost the same accuracy, but with half the average bit budget. Although this may seem counter-intuitive at the first glance, we remark that due to random batch sampling in SGD, gradient values behave as random variables, and hence, using model difference over $H$ iterations may provide a more accurate observation to be able to identify new important weights. To reduce the bit budget further we consider TCS-L4-Q5, where all the non-zero values are represented with 5 bits in total while one bit is used for the sign. We observe that TCS with quantization can achieve a $\times2000$ reduction in the communication load, with an even better test accuracy compared to the centralized baseline.

\begin{table}
\begin{center}
\begin{tabular}{ |c | c| c|}
    \hline
    \textbf{Method} & \textbf{Test Accuracy (mean $\pm$ std)} & \textbf{Bit budget}\\ 
    \hline
    TCS & 92.094 $\pm$ 0.182 & 0.363\\
    \hline
    TCS-L4 &92.62$\pm$ 0.0892& 0.0907\\
    \hline
    TCS-L4-Q5 &\textbf{92.763 $\pm$ 0.210}& 0.01675\\
    \hline
    TCS-LF &92.481 $\pm$ 0.157& 0.363\\
    \hline
    TCS-PLF & 92.442 $\pm$ 0.189 & 0.363\\
    \hline
    TCS-LF-Q5 &92.042 $\pm$ 0.173& 0.067\\
    \hline
    top-$K$ & 91.856 $\pm$ 0.317 & 0.41\\
    \hline
    Baseline & 92.228 $\pm$ 0.232 & -\\
    \hline
\end{tabular}
\end{center}
 \caption{Test accuracy  (for $\eta=0.8$) and bit budget of the studied schemes.}
 \label{top1_acc_08}
\end{table}

We observe that when $\eta=0.5$, layer-wise fairness (TCS-LF) does not improve the accuracy. We argue that the correlation over time may not be identical for all the layers, which requires more custom choice for the fairness constraints. We also remark that the efficiency of TCS and TCS-LF may depend on the learning rate. To analyze the impact of the learning rate $\eta$ on the performance of TCS and its variations with layer-wise fairness, we repeat our simulations with a learning rate of $\eta=0.8$ and report the results in Table \ref{top1_acc_08}. The results show that, when the learning rate is increased, although the test accuracy of the TCS slightly reduces, its layer-wise fair variations TCS-LF and TCS-PLF perform better. Similarly to the previous simulation results with $\eta=0.5$, we observe that the highest test accuracy is achieved when TCS is implemented with multiple local steps.\\
\indent We emphasize that the impact of quantization on the bit budget is more visible  with TCS compared to top-$K$ sparsification. When quantization is used with top-$K$ sparsification, the  number of bits used for the location becomes the bottleneck as quantization cannot reduce that. When TCS (with $\phi_{global} = 10^{-2}$ and $\phi_{local} = 10^{-3}$) is employed together with $5$-bit quantization, the corresponding bit budget is $0.067$ (bits per element). On the other hand, top-$K$ sparsification ($\phi = 10^{-2}$) with $5$-bit quantization requires a bit budget of $0.14$, which is more than twice the bit budget of TCS. Furthermore, when the number of bits used for quantization decreases, TCS becomes more and more communication efficient. 

\begin{figure*}[t]
\centering
\includegraphics[scale =1.0]{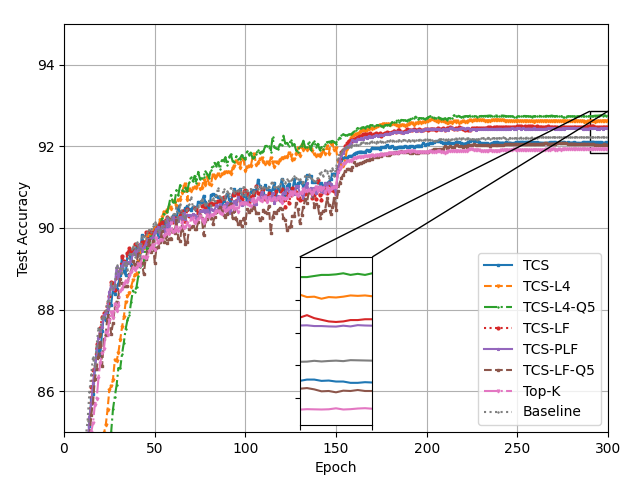}
\caption{Comparison of the test accuracy results for different FL techniques and the centralized baseline over 300 epochs (for $\eta=0.8$).}
\label{300epoch_08}
\end{figure*}

\begin{figure*}[h]
\centering
\includegraphics[scale =1.0]{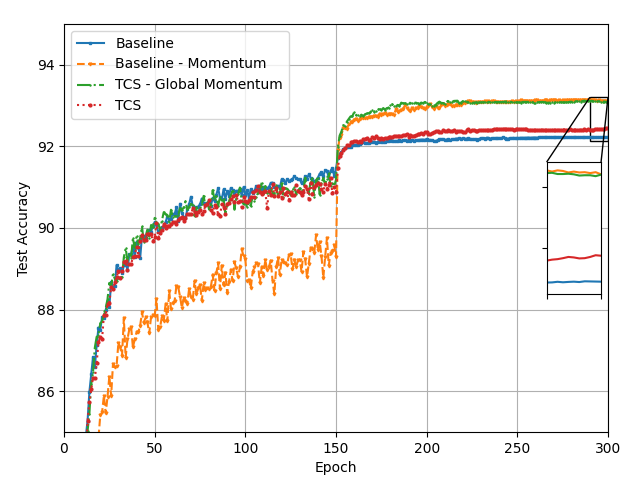}
\caption{Comparison of the baseline and TCS strategies with and without momentum.}
\label{300_epoch grad}
\end{figure*}

Finally, we evaluate the performance of TCS with global momentum, introduced in Section \ref{sec:TCS-m} with parameter $\beta=0.9$, and compare it with the centralized baseline with momentum SGD. The final test accuracy results are presented in Table \ref{top1_acc_grad} and illustrated in Figure \ref{300_epoch grad}. The results show that TCS with global momentum achieves the same test accuracy result with the centralized baseline while providing $\times100$ reduction in the communication load as before. Hence, we can conclude that momentum can be efficiently incorporated into the TCS framework.

\begin{table}
\begin{center}
\begin{tabular}{ |c | c| c|}
    \hline
    \textbf{Method} & \textbf{Test Acc.}\\ 
    \hline
    Baseline & 92.23 $\pm$ 0.232 \\
    \hline
    Baseline-Momentum &\textbf{93.105 $\pm$ 0.346}\\
    \hline
    TCS- Global Momentum& 93.102 $\pm$ 0.182 \\
    \hline
    TCS &92.44 $\pm$ 0.128 \\
    \hline
\end{tabular}
\end{center}
\caption{Test accuracy results of  Baseline, Baseline with momentum, TCS with global m omentum and TCS.}
 \label{top1_acc_grad}
\end{table}

\section{Conclusion}
In this paper, we introduced a novel sparse communication strategy for communication-efficient FL, called time correlated sparsification (TCS), by establishing an analogy between network pruning and gradient sparsification frameworks. The proposed strategy is built upon the assumption that at ``important'' locations the model difference (or the gradient) changes slowly over time, and utilizes this correlation over iterations to reduce the communication load. Through extensive simulations on CIFAR-10 dataset, we show that TCS can meet or even surpass the centralized baseline accuracy with $\times100$ sparsification, and can reach up to $\times2000$ reduction in the communication load when it is employed together with quantization. The proposed TCS strategy provides natural sparsification in the downlink communication as well. By employing a separate quantization operator at the PS, similarly to \cite{SGD.feedback_double1, SGD.feedback_double2, Amiri:FedQuantPS}, the communication load for the downlink phase can be reduced further, which we will investigated in an  extension of this work.

\bibliographystyle{IEEEtran}
\bibliography{IEEEabrv,ref.bib}

\ifCLASSOPTIONcaptionsoff
  \newpage
\fi

\end{document}